\documentclass{article}
\usepackage[utf8]{inputenc}
\usepackage{geometry}
\usepackage{soul}

\usepackage{cite}
\usepackage{graphicx}

\usepackage{mathpazo}
\usepackage{textcomp}

\usepackage{authblk}

\makeatletter
\newcommand*\titleheader[1]{\gdef\@titleheader{#1}}
\AtBeginDocument{%
  \let\st@red@title\@title
  \def\@title{%
    \bgroup\normalfont\small\centering\@titleheader\par\egroup
    \vskip1.5em\st@red@title}
}
\makeatother

\title{Scaling of a large-scale simulation of synchronous slow-wave and asynchronous awake-like activity of a cortical model with long-range interconnections} 
\titleheader{Final peer reviewed version as: E. Pastorelli et al. "Scaling of a Large-Scale Simulation of Synchronous Slow-Wave and Asynchronous Awake-Like Activity of a Cortical Model With Long-Range Interconnections", Front Syst Neurosci. 2019;13:33. Published 2019 Jul 23. doi:10.3389/fnsys.2019.00033}

\usepackage{authblk}
\author[1,2,*] {Elena Pastorelli}
\author[1] {Cristiano Capone}
\author[1] {Francesco Simula}
\author[4,5] {Maria V. Sanchez-Vives} 
\author[1,3] {Paolo Del Giudice} 
\author[3] {Maurizio Mattia} 
\author[1] {Pier Stanislao Paolucci}

\affil[1]{\small INFN, Sezione di Roma, Rome, Italy}
\affil[2]{\small PhD Program in Behavioural Neuroscience, “Sapienza” University, Rome, Italy}
\affil[3]{\small Istituto Superiore di Sanità, Rome, Italy}
\affil[4]{\small Systems Neuroscience, IDIBAPS, Barcelona, Spain}
\affil[5]{\small ICREA, Barcelona, Spain}
\affil[*]{\small Corresponding author: Elena Pastorelli, elena.pastorelli@roma1.infn.it}

\begin{document}
\date{}
\maketitle

\begin{abstract}

Cortical synapse organization supports a range of dynamic states on multiple spatial and temporal scales, from synchronous slow wave activity (SWA), characteristic of deep sleep or anesthesia, to fluctuating, asynchronous activity during wakefulness (AW). Such dynamic diversity poses a challenge for producing efficient large-scale simulations that embody realistic metaphors of short- and long-range synaptic connectivity. In fact, during SWA and AW different spatial extents of the cortical tissue are active in a given timespan and at different firing rates, which implies a wide variety of loads of local computation and communication. A balanced evaluation of simulation performance and robustness should therefore include tests of a variety of cortical dynamic states. Here, we demonstrate performance scaling of our proprietary Distributed and Plastic Spiking Neural Networks (DPSNN) simulation engine in both SWA and AW for bidimensional grids of neural populations, which reflects the modular organization of the cortex. We explored networks up to 192$\times$192 modules, each composed of 1250 integrate-and-fire neurons with spike-frequency adaptation, and exponentially decaying inter-modular synaptic connectivity with varying spatial decay constant. For the largest networks the total number of synapses was over 70 billion. The execution platform included up to 64 dual-socket nodes, each socket mounting 8 Intel Xeon Haswell processor cores @ 2.40GHz clock rate. Network initialization time, memory usage, and execution time showed good scaling performances from 1 to 1024 processes, implemented using the standard Message Passing Interface (MPI) protocol. We achieved simulation speeds of between 2.3$\times$$10^9$ and 4.1$\times$$10^9$ synaptic events per second for both cortical states in the explored range of inter-modular interconnections.

\end{abstract}

\section{Introduction}
At the large scale, the neural dynamics of the cerebral cortex result from an interplay between local excitability and the pattern of synaptic connectivity. This interplay results in the propagation of neural activity. A case in point is the spontaneous onset and slow propagation of low-frequency activity waves during the deep stages of natural sleep or deep anesthesia \cite{reyes-puerta:2016,destexhe2011,hobson2002,sanchez2014slow}.

The brain in deep sleep expresses slow oscillations of activity at the single-neuron and local network levels which, at a macroscopic scale, appear to be synchronized in space and time as traveling waves (slow-wave activity, SWA). The “dynamic simplicity” of SWA is increasingly being recognized as an ideal test bed for refining and calibrating network models composed of spiking neurons. Understanding the dynamical and architectural determinants of SWA serves as an experimentally grounded starting point to tackle models of behaviorally relevant, awake states \cite{Han2008,Luczak2009,Curto2009}. A critical juncture in such a logical sequence is the description of the dynamic transition between SWA and asynchronous, irregular activity (AW, asynchronous wake state) as observed  during fade-out of anesthesia, for instance, the mechanism of which is still a partially open problem \cite{solovey2015loss,Curto2009,Steyn2013}. To help determine the mechanism of this transition, it may be of interest to identify the factors enabling the same nervous tissue to express global activity regimes as diverse as SWA and AW. Understanding this repertoire of global dynamics requires high-resolution numerical simulations of large-scale networks of neurons which, while keeping a manageable level of simplification, should be realistic with respect to both nonlinear excitable local dynamics and to the spatial dependence of the synaptic connectivity (as well as the layered structure of the cortex) \cite{Hill2005,Bazhenov2002,potjans:2014,Krishnan2016}.

Notably, efficient brain simulation is not only a scientific tool, but also a source of requirements and architectural inspiration for future parallel/distributed computing architectures, as well as a coding challenge on existing platforms. Neural network simulation engine projects have focused on: flexibility and user friendliness, biological plausibility, speed and scalability (e.g. NEST \cite{gewaltig:2007,jordan:2018}, NEURON \cite{hines:1997,carnevale:2006}, GENESIS \cite{NIPS1988_182}, BRIAN \cite{goodman:2009,stimberg:2014}). Their target execution platforms can be either homogeneous or heterogeneous (e.g. GPGPU-accelerated) high-performance computing (HPC) systems, \cite{modha:2011,izhikevich:2008,nageswaran:2009}, or neuromorphic platforms, for either research or application purposes (e.g. SpiNNaker \cite{furber:2012}, BrainScaleS \cite{schmitt:2017}, TrueNorth \cite{merolla:2014}).

From a computational point of view, SWA and AW pose different challenges to simulation engines, and comparing the simulator performance in both situations is an important element in assessing the general value of the choices made in the code design. During SWA, different and limited portions of the network are sequentially active, with a locally high rate of exchanged spikes, while the rest of the system is almost silent. On the other hand, during AW the whole network is homogeneously involved in lower rate asynchronous activity. In a distributed and parallel simulation framework, this raises the question of whether the computational load on each core and the inter-process communication traffic are limiting factors in either cases. We also need to consider that activity propagates for long distances across the modeled cortical patch, therefore the impact of spike delivery on the execution time depends on the chosen connectivity. Achieving a fast and flexible simulator, in the face of the above issues, is the purpose of our Distributed and Plastic Spiking Neural Networks (DPSNN) engine. Early versions of the simulator~\cite{paolucci_dpsnn:2013} originated from the need for a representative benchmark developed to support the hardware/software co-design of distributed and parallel neural simulators. DPSNN was then extended to incorporate the event-driven approach of \cite{mattia:2000}, implementing a mixed time-driven and event-driven strategy similar to the one introduced in \cite{morrison:2005}. Here, we report the performances of DPSNN in both slow-wave (SW) and AW states, for different sizes of the network and for different connectivity ranges. Specifically, we discuss network initialization time, memory usage and execution times, and their strong and weak scaling behavior. 

\section{Materials and methods}
\label{sec:materials}
\subsection{Description of the Distributed Simulator}
\label{sec:simDesc}
DPSNN has been designed to be natively distributed and parallel, with full parallelism also exploited during the creation and initialization of the network. The full neural system is represented in DPSNN by a network of C++ processes equipped with an agnostic communication infrastructure, designed to be easily interfaced with both Message Passing Interface (MPI) and other (custom) software/hardware communication systems. Each C++ process simulates the activity of one or more clusters of neighboring neurons and their set of incoming synapses. Neural activity generates spikes with target synapses in other processes; the set of “axonal spikes” is the payload of the associated exchanged messages. Each axonal spike carries the identity of its producing neuron and its original time of emission (AER, Address Event Representation \cite{lazzaro:1993}). Axonal spikes are only sent to target processes where at least one target synapse exists.

The memory cost of point-like neuron simulation is dominated by the representation of recurrent synapses which, in the intended biologically plausible simulations, are numbered in thousands per neuron. When plasticity support is switched off, the local description of each synapse includes only the identity of the target and source neurons, the synaptic weight, the transmission delay from the pre- to the post-synaptic neuron, and an additional optional identifier for possible different types of synapse (see Table~\ref{tab:synrepresent}).

\begin{table}[hbt]
\small
\caption{Representation of recurrent synapses. Static synapses cost 12 bytes per synapse. Plasticity support adds a cost of 8 bytes per synapse}
\label{tab:synrepresent}
\begin{center}
\begin{tabular}{|c|c|c|c|c|c|c|c|}
\hline
\multicolumn{8}{|c|}{Synaptic representation} \\
\hline
\multicolumn{6}{|c|}{Static (12 bytes/synapse)} & \multicolumn{2}{|c|}{Plasticity (8 bytes/synapse)}\\
\hline
FIELD & Source neuron ID & Target neuron ID & Weight & Delay & Kind & Last spiking time & Derivative\\
\hline
SIZE (byte) & 4 & 4 & 2 & 1 & 1 & 4 & 4\\
\hline
\end{tabular}
\end{center}
\end{table}

When the synaptic plasticity support is switched on, each synapse takes note of its own previous activation time.
For every postsynaptic neuron spike and synaptic event a Spike-timing-dependent plasticity (STDP) contribution is computed using double floating point precision. Individual STDP events contribute to synaptic Long Term Potentiation (LTP) and Depression (LTD) through a first order low pass filter operating at a longer timescale than that of neural dynamics. Intermediate computations are performed in double precision, while the status of the low-pass filter is stored in the synaptic representation using single floating point precision.
In the present work, synaptic plasticity is kept off.

During the initialization phase only, each synapse is represented at both the processes storing the source and target neuron because, in the current implementation, each process sends to each member in the subset of processes to be connected a single message requesting the creation of all needed synaptic connections. This memory overhead could be reduced by splitting the generation of connection requests in group, at the price of a proportional increase in the number of communication messages needed by the initialization phase. Indeed this is the moment of peak memory usage. Afterwards, the initialization synapses are stored exclusively in the process-hosting target neurons. For each process, the list of all incoming synapses is maintained. The synaptic list is double-ordered: synapses with the same delay are grouped together and are further ordered by presynaptic neuron index as in \cite{mattia:2000}. Incoming spikes are ordered according to the identity of the presynaptic neurons. The ordering of both the synaptic list and incoming spikes helps a faster execution of the demultiplexing stage that translates incoming spikes into synaptic events because it increases the probability of contiguous memory accesses while exploring the synaptic list.

Because of the high number of synapses per neuron the relative cost of storing the neuron data structure is relatively small. It mainly contains the parameters and status variables used to describe the dynamics of the single-compartment point-like neuron. Moreover, it also contains the queue of input spikes (both recurrent and external) that arrived during the previous simulation time step, with their associated post-synaptic current value and time of arrival. External spikes are generated as a Poissonian train of synaptic inputs. In the neuron data structure, double-precision floating-point storage is adopted for the variables used in the calculation of the membrane potential dynamics, while all the other variables and constant parameters related to the neuron are stored using single precision. Also, we opted for double precision floating point computations, because of the dominance of the cost related to the transport and memory accesses while distributing neural spikes to synapses.

In DPSNN there is no memory structure associated with external synapses: for each neuron, for each incoming external spike the associated synaptic current is generated on the fly from a Gaussian distribution with assigned mean and variance (see Section \ref{sec:gridAndStates}); therefore external spikes have a computational cost but a negligible memory cost. The described queuing system ensures that the full set of synaptic inputs, recurrent plus external, are processed using an event-driven approach.

\subsubsection{Execution Flow: Overview of the Mixed Time- and Event-driven Simulation Scheme}
There are two phases in a DPSNN simulation: (1) the creation of the structure of the neural network and its initialization; (2) the simulation of the dynamics of neurons (and synapses, if plasticity is switched on). For the simulation phase we adopted a combined event-driven and time-driven approach partially inspired by \cite{morrison:2005}. Synaptic events drive the neural dynamics, while the message passing of spikes among processes happens at regular time steps, which must be set shorter than the minimum synaptic delay to guarantee the correct causal relationship in the distributed simulation.   The minimum axo-synaptic delay sets the communication time step of the simulation, in our case 1ms.

\begin{figure}[!hbt]
\centering
\includegraphics[width=0.80\textwidth]{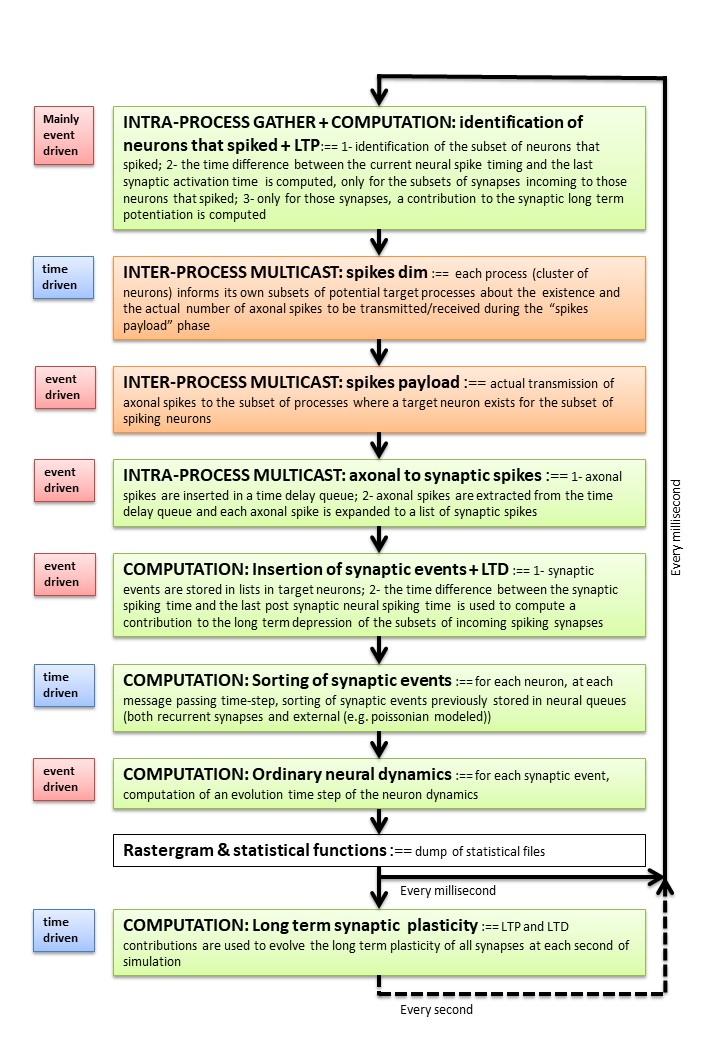}
\caption{Execution flow. Each software process composing a DPSNN simulation represents one (or more) clusters of spatially neighboring neurons and their incoming synapses. Each process iterates over the blocks listed here which simulates the activity of local neurons and incoming synapses, their plasticity and the exchange of messages through axo-dendritic arborizations. It is a sequential flow of event-driven and time-driven computational and communication (inter- and intra-process) blocks (see the label on the left of each block). The measure of the relative execution times of the blocks is used to guide  the optimization effort and to drive the co-design of dedicated hardware/software architectures.}
\label{fig:flow}
\end{figure}

Figure~\ref{fig:flow} describes the main blocks composing the execution flow and the event- or time-driven nature of each block.
The simulation phase can be broken down into the following phases: (1) identification of the subset of neurons that spiked during the previous time step, and (when plasticity is switched on) computation of an event-driven STDP contribution; (2) spikes are sent to the cluster of neurons where target synapses exist (inter-process communication blocks in the figure); (3) the list of incoming spikes to each process is placed into the double-ordered synaptic list, waiting for a number of time steps that match the synaptic delays, at which point the corresponding target synapses are activated; (4) synapses inject their event into queues that are local to their post-synaptic neuron and compute the STDP plasticity contribution; (5) each neuron sorts the lists of input events produced by recurrent and external synapses; (6) for each event in the queue, the neuron integrates its own dynamic equations using an event-driven solver. Periodically, at a slower time step (1~s in the current implementation), all  synapses modify their efficacies using the integrated plasticity signal described above. Later sections describe the individual stages and data representation in further detail.

\subsubsection{Spike Messages: Representation and Communication}
Spike messages are defined according to an AER, where each spike is represented by the identity of the spiking neuron, and the spiking time. Spikes sharing a target process (i.e., targeting the cluster of neurons managed by a process) are packed into “axonal spikes” messages that are delivered from the source process to its target processes during the communication phase. During execution, the “axon” arborization is managed by the target process in order to reduce the network communication load. Its construction during the initialization phase involves the following steps: in a “fan-out” step, each process notifies the processes which will host the target synapses of its own neurons. In the next “fan-in” step, each target process stores its own list of source neurons, and the associated double-ordered list of incoming synapses. For details about the initial construction of the connectivity infrastructure and the delivery of spiking messages see Supplementary Materials.

\subsubsection{Bidimensional Grids of Cortical Columns and their Mapping onto Processes}
Neural networks are organized in this study as bidimensional grids of modules (mimicking cortical columns) as in \cite{capone:2019}. Each module is composed of 1250 point-like neurons, and further organized into subpopulations. The set of cortical columns can be distributed over a set of processes (and processors): each process can either host a fraction of a column (e.g. \textonehalf, \textonequarter…), a whole single column, or several columns (up to 576 columns per process for the experiments reported in this paper). In this respect, our data distribution strategy aims to store in contiguous memories neighboring neurons and their incoming synapses. As usual in parallel processing, for a given problem size the performance tends to saturate and then to worsen, as it is mapped onto an excessively large number of processors, so that for each grid size an optimal partitioning onto processes is heuristically established, in terms of the performance measure defined in Section \ref{sec:perfMeasures}.

Each process runs on a single processor; if hyperthreading is active, more than one process can be simultaneously executed on the same processor. For the measures in this paper, hyperthreading was deactivated, so that one hardware core always runs a single process. In DPSNN, distributing the simulation over a set of processes means that both the initialization and the run phases are distributed, allowing the scaling of both phases with the number of processes, as demonstrated by the measures reported in Section~\ref{sec:results}.

\subsection{Model Architecture and Network States}
\label{sec:gridAndStates}
For each population in the modular network, specific parameters for the neural dynamics can be defined, as well as specific intra- and inter-columnar connectivity and synaptic efficacies. Connectivity among different populations can be modeled with specific laws based on distance-dependent probability, specific to each pair of source and target subpopulation.  By suitably setting the available interconnections between different populations, cortical laminar structures can also be potentially modeled in the simulation engine. 

The grids (see Table \ref{tab:configs}) are squares in a range of sizes (24$\times$24, 48$\times$48, 96$\times$96, 192$\times$192). Each local module is always composed of $K=1250$ neurons, further subdivided into subpopulations. We implemented a ratio of 4:1 between excitatory and inhibitory neurons ($K_{I}$=250 neuron/module); in each module, excitatory neurons ($E$) were divided into two subpopulations: $K_{F}$=250 (25\%) strongly coupled “foreground” neurons ($F$), having a leading role in the dynamics, and $K_{B}$=750 (75\%) “background” neurons ($B$) continuously firing at a relatively low rate. Populations on the grid are connected to each other through a spatial connectivity kernel. The probability of connection from excitatory neurons decreases exponentially with the inter-module distance $d$:
\begin{equation}
C_{ts\lambda}(d) = C_{ts\lambda}^0 \times exp(\frac{-d}{\lambda}) \, .
\label{eq:1}
\end{equation}
More specifically, $d$ is the distance between the source ($s=\{F,B\}$), and target ($t=\{F,B,I\}$) module, and $\lambda$ a characteristic spatial scale of connectivity decay. $d$ and $\lambda$ are expressed using inter-modular distance units (imd).  For the simulations here described, the translation to physical units sets imd in the range of a few hundreds micrometers. Simulations are performed considering different $\lambda$ values (0.4, 0.5, 0.6, 0.7) imd, but $C_{ts\lambda}^0$ is set so as to generate the same mean number of projected synapses per neuron ($M_{t}=0.9*K_{t}, t=\{F,B,I\}$) for all $\lambda$ values. Connections originating from inhibitory neurons are local (within the same local module) and also in this case $M_{t}=0.9*K_{t}, t=\{F,B,I\}$.  All neurons of the same type (excitatory/inhibitory) in a population share the same mean number of incoming synapses. The connectivity has open boundary conditions on the edges of the two-dimensional surface.

Synaptic efficacies are randomly chosen from a Gaussian distribution with mean $J_{ts}$ and SD $\Delta J_{ts}$, chosen in different experiments so as to set the system in different working regimes and simulate different states. The procedure for the selection of the efficacies is based on a mean-field method described below. Each neuron also receives spikes coming from neurons belonging to virtual external populations, collectively modeled as a Poisson process with average spike frequency $\nu_{ext}$ and synaptic efficacy $J_{ext}$. Excitatory neurons are point-like leaky integrate-and-fire (LIF) neurons with spike frequency adaptation (SFA) \cite{gigante:2007,capone:2019}. SFA is modeled as an activity-dependent self-inhibition, described by the fatigue variable $c(t)$. The time evolution of the membrane potential $V(t)$, and $c(t)$, of excitatory neurons between spikes is governed by:
\begin{equation}
\label{eq:2}
\left\{
\begin{array}{lll}	
\dot{V} &= &-\frac{V-E}{\tau_{m}} - g_{c} \frac{c}{C_{m}} + \sum J_{i}\delta(t-t_{i}-\delta_{i}) + \sum J_{ext}^i \delta(t-t_{i}^{poiss})
\\[12pt]
\dot{c} &= &-\frac{c}{\tau_{c}} + \alpha_{c} \, \sum_{k} \delta(t-t^{(k)}_{\mathrm{sp}})
\end{array}\right.
\end{equation}
$\tau_m$ is the membrane characteristic time, $C_{m}$ the membrane capacitance, and $E$ the resting potential. SFA is not considered for inhibitory neurons; that is, in (\ref{eq:2}), the second equation and the $g_{c} \frac{c}{C_{m}}$ term in the first are absent. Incoming spikes, generated at times $t_{i}$, reached the target neuron with delay $\delta_{i}$ and provoked instantaneous membrane potential changes of amplitude $J_{i}$. Alike, external stimuli produce a $J_{ext}^i$ increment in the membrane potential, with $t_{i}^{poiss}$ representing the spike times generated by a Poisson distribution of average $\nu_{ext}$. An output spike at time $t^{(k)}_{\mathrm{sp}}$ was triggered if the membrane potential exceeded a threshold $V_{\theta}$. On firing, the membrane potential was reset to $V_{r}$ for a refractory period $\tau_{arp}$, whereas $c$ was increased by the amount $\alpha_{c}$. Once the network connectivity and the neural dynamics have been defined, synaptic efficacies and external stimuli can be set to determine the dynamical states accessible to the system, by means of mean-field theory. 

We elaborated a dynamical mean-field description for our simulations.  We assume that the input received by the different neurons in a module are independent but are conditioned by the same (and possibly time-dependent) mean and variance of the input synaptic current \cite{brunel1999fast,mattia2002population}.

The gain function $\phi_{i}$ for this kind of neurons was firstly found by \cite{ricciardi:1977} as the first-passage time (FPT) of the membrane potential to the firing threshold, for an integrate-and-fire neuron with stationary white noise input current, in the diffusion approximation.

Spike frequency adaptation is an important ingredient for the occurrence of slow oscillations in the mean field dynamics \cite{gigante:2007,capone:2019}.

The mean-field dynamics for the average activity $\nu_{i}, i = \{F,B,I\}$ is determined by the gain functions $\phi_{i}$ as follows:
\begin{equation}
\label{eq:3}
(F,B): \left\{
\begin{array}{lll}	
\dot{\nu_{i}} &= &\frac{\phi_{i}(\overrightarrow{\nu},\overrightarrow{c})-\nu_{i}}{\tau_{E}}
\\ [12pt]
\dot{c}_{i} &= &-\frac{c_{i}}{\tau_{c}} + \alpha_{c}\nu_{i}
\end{array}\right.
\,; \,\,\,\,\,\,\,\,\,
(I): \,\,	
\dot{\nu_{i}} = \frac{\phi_{i}(\overrightarrow{\nu})-\nu_{i}}{\tau_{I}}
\end{equation}
where $\tau_{E}$ and $\tau_{I}$ are phenomenological time constants. 
The interplay between the recurrent excitation embodied in the gain function, and the activity-dependent self-inhibition, is the main driver of the alternation between a high-activity (Up) state and a low-activity (Down) state \cite{mattia:2012,capone:2019}.

In the mean-field description of the neural module, synaptic connectivity is chosen so as to match the total average synaptic input the neurons would receive inside the multi-modular network. The fixed points of the dynamics expressed by equations (\ref{eq:3}) can be analyzed using standard techniques \cite{strogatz:2018}. The nullclines of the system (where $\dot{\nu}=0$ or $\dot{c}=0$) cross at fixed points that can be predicted to be either stable or unstable. In the simulations described here, the strengths of recurrent synapses $J_{t,s}$ connecting source population $s$ and target population $t$, and of external synapses $J_{t,ext}$, is randomly chosen from a Gaussian distribution with mean $J_{t,s}$ and variance $\Delta$$J_{t,s} = 0.25 \times J_{t,s}$.

\begin{figure}[!hbt]
\centering
\includegraphics[width=1.0\textwidth]{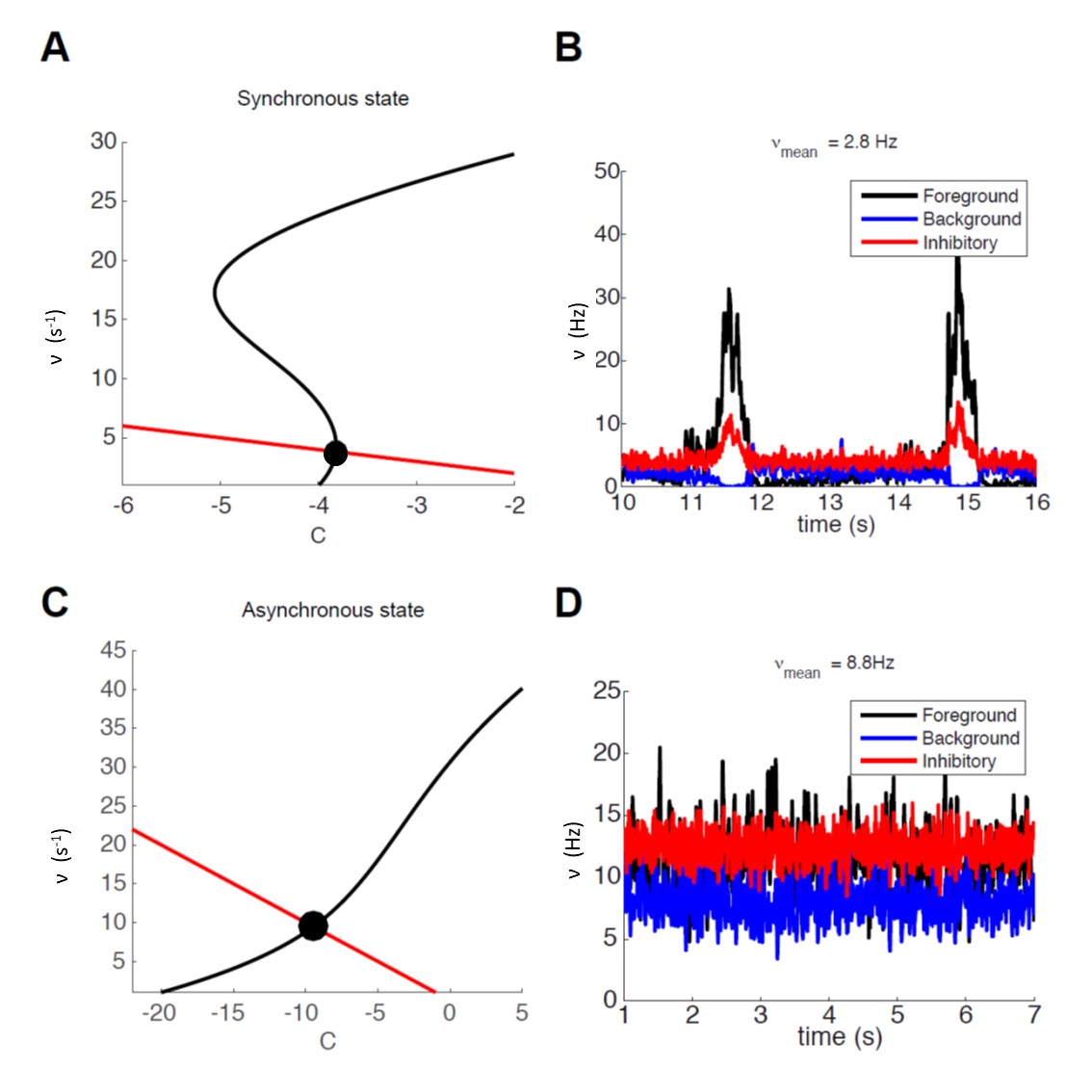}
\caption{Dynamical representation of the SW and AW states. \textbf{(A)} Phase space representation of mean-field analysis with a weakly stable fixed point. \textbf{(B)}  Firing-rate time course for an example module, for foreground, background, and inhibitory sub-populations (respectively in black, blue, and red) in the SW state. \textbf{(C)} Phase space representation of mean-field analysis with a stable fixed point at a high level of activity. \textbf{(D)}  Firing-rate time course for an example module, for foreground, background, and inhibitory sub-populations (respectively in black, blue, and red) in the AW state.}
\label{fig:dynStates1}
\end{figure}

We relied on mean-field analysis to identify neural and network parameters setting the network's modules in SW or AW dynamic regimes.  
Figure~\ref{fig:dynStates1}A shows an example of nullclines for the mean-field equations (\ref{eq:3}) of a system displaying SW. The black S-shaped line is the nullcline for the rate $\nu$  while the red straight line is the one for the fatigue variable $c$  (for details see \cite{mattia:2012,capone:2019}). The stable fixed point, at the intersection of the nullclines, has a low level of activity and is characterized by a small basin of attraction: the system can easily escape from it thanks to the noise, and it gets driven towards the upper branch of the $\nu$ nullcline from which, due to fatigue, it is attracted back to the fixed point, thus generating an oscillation (see Figure~\ref{fig:dynStates1}B).

Network parameters can also be set in order to have an asynchronous state, mainly by setting a lower Foreground-to-Foreground (FF) synaptic efficacy, which generates a more linear $\nu$ nullcline close to the fixed point (see Figure~\ref{fig:dynStates1}C). In this case the basin of attraction of the fixed point is larger, and oscillations do not occur, resulting in a stationary asynchronous state (see Figure~\ref{fig:dynStates1}D), in which neural activity fluctuates around the mean-field fixed point. Table~S1 in Supplementary Materials reports the complete list of synaptic parameters for both SW and AW states.

\subsection{Neural Activity Analysis}
The simulation generates the spikes produced by each neuron in the network. From these, the time course of the average firing rate for each subpopulation in the network can be computed using an arbitrary sampling step, which here we set to be  5~ms. Power spectra are computed using the Welch method (see Figures~\ref{fig:scaling}B and~\ref{fig:scaling}E for examples of SW and AW power spectra including delta band).

A proxy to experimentally acquired multi-unit activity ($\mathrm{MUA(t)}$) is computed as the average firing rate of a simulated module.
In similarity with experimental data analysis \cite{capone:2019} we considered the logarithm of such signal ($\log \frac{ \mathrm{MUA(t)} }{\mathrm{MUA}_{down}}$) where the $\mathrm{MUA}_{down}$ is the average $\mathrm{MUA(t)}$ in the Down state. 
A zero mean white noise is added to emulate unspecific background fluctuations from neurons surrounding the module. The variance of such noise ($0.5$) was chosen to match the width of experimental $\log \frac{\mathrm{MUA (t)}}{\mathrm{MUA}_{down}}$ distributions for bistable neural populations in their Down state.

For SW, the $\mathrm{MUA}(t) $ signal alternated between high and low activity states (Up and Down states). $T(x,y)$, the Down-to-Up transition time, is determined by a suitable threshold and is detected for each point in the spatial grid, defining the propagation wavefront. $V(x,y)$, the absolute value of wavefront speed, can be computed as  
\begin{equation}
\label{eq:4}
V(x,y) = \frac{1}{\sqrt{(\frac{\partial T(x,y)}{\partial x})^2 + (\frac{\partial T(x,y)}{\partial y})^2}}
\end{equation}
The average speed during the collective Down-to-Up transitions is obtained by averaging $V(x,y)$ over all the simulated positions $(x,y)$.

\subsection{Performance Measures}
\label{sec:perfMeasures}
Performance is quantified in terms of “strong scaling” and “weak scaling”: the former refers to keeping the problem size fixed and varying the number of hardware cores, while the latter refers to the increase in equal proportions of the problem size and the number of hardware cores.

As a performance measure we computed the equivalent synaptic events per second as the product of the total number of synapses (recurrent and external) and the number of spikes occurred across the whole simulation, divided by the elapsed execution time. This way, a comparison of the simulation cost among different problem sizes and hardware/software resources (core/processes) can be captured in a single graph. Similarly, we defined a convenient metric to evaluate the memory efficiency of a simulation: by dividing the total memory required by the simulation by the number of recurrent synapses to be represented. Indeed, as stated in Section \ref{sec:simDesc} we expect the memory usage to be dominated by the representation of synapses which are thousands per neuron. 

Scalability measurements were taken on different neural network sizes, varying the size of the grid of columns and, for each size, distributing it over a different span of MPI processes. We selected four grid sizes: 24$\times$24 columns, including 0.7M neurons and 1.1G synapses; 48$\times$48 columns including 2.8M neurons and 4.4G synapses; 96$\times$96 columns including 12M neurons and 17.6G synapses; 192$\times$192 columns including 46M neurons and 70G synapses. The number of processes over which each network size was distributed varied from a minimum, bounded by memory, and a maximum, bounded by communication or HPC platform constraints (see Table~\ref{tab:configs}). For the 192$\times$192 configuration, only one measure was taken because of memory requirements, corresponding to a run distributed over 1024 MPI processes/hardware cores. 

Execution times for SW were measured across the time elapsed between the rising edges of two subsequent waves, and for AW across a time span of 3~s. In both cases, initial transients were omitted.

\begin{table}[htb]
\small
\caption{Configurations used for the scaling measures of DPSNN.}
\label{tab:configs}
\begin{center}
\begin{tabular}{|c|c|c|c|c|c|c|}
\hline
Grid & Number of columns & Number of neurons & \multicolumn{2}{|c|}{Number of synapses} & \multicolumn{2}{|c|}{MPI proc}\\
\hline
{ } & { } & { } & Recurrent &  Total &  Min &  Max\\
\hline
24x24 & 576 & 0.72M & 0.8G & 1.1G & 1 & 64\\
\hline
48x48 & 2304 & 2.88M & 3.2G & 4.4G & 4 & 256\\
\hline
96x96 & 9216 & 11.52M & 12.9G & 17.6G & 64 & 1024\\
\hline
192x192 & 36864 & 46.08M & 51.8G & 70.3G & 1024 & 1024\\
\hline
\end{tabular}
\end{center}
\end{table}

\subsection{Validation of Results and Comparison of Performances}
\label{sec:DPSNNvsNESTmaterial}
\begin{figure}[!hbt]
\centering
\includegraphics[width=1.0\textwidth]{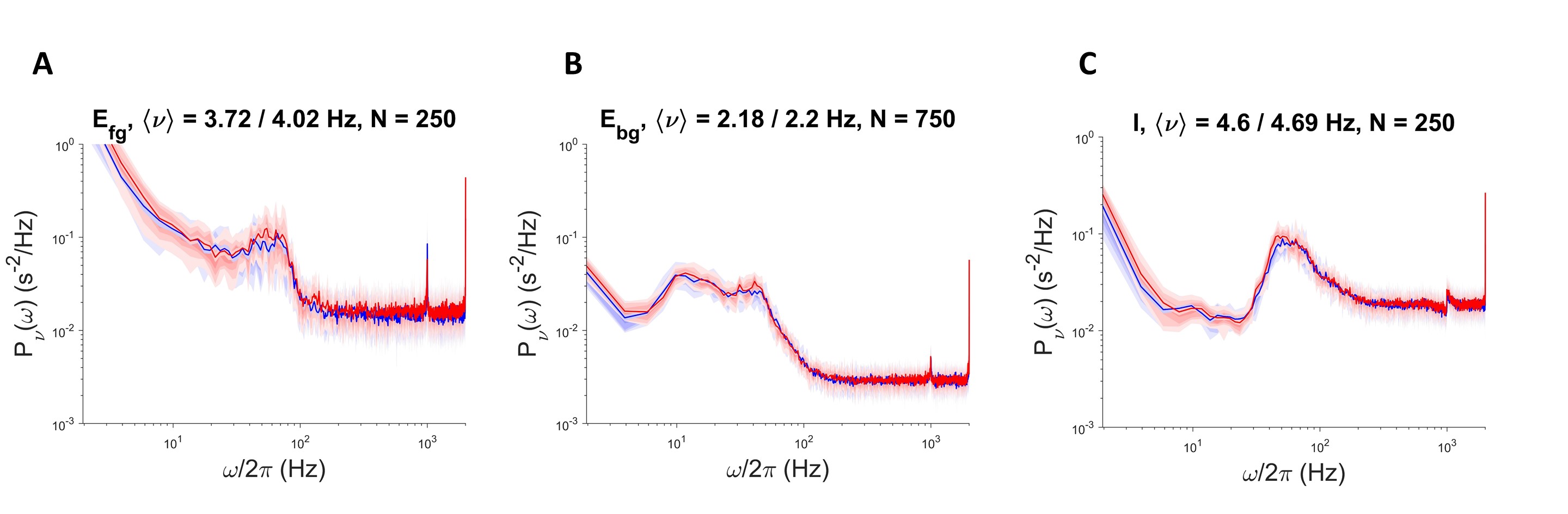}
\caption{PSD comparison of NEST (red) and DPSNN (blue) simulation output. Comparison performed on a single sub-population located in a specific cortical column of the simulated neural network grid. From left to right, PSD of foreground excitatory \textbf{(A)}, background excitatory \textbf{(B)} and inhibitory \textbf{(C)} neural subpopulations.}
\label{fig:PSDcomparison}
\end{figure}

We used NEST version 2.12, the high-performance general purpose simulator developed by the NEST Initiative, as a reference for the validation of results produced by the specialized DPSNN engine and for the comparison of its performances (speed, initialization time, memory footprint). 
The comparison was performed for both SW and AW dynamical states, for a network of 4.4G synapses (48$\times$48 grid of columns). We assessed the correctness of results using power spectral density (PSD) analysis of the temporal evolution of the firing rates of each subpopulation. Figure~\ref{fig:PSDcomparison} reports an example of this comparison, showing the PSD match of NEST and DPSNN for each single subpopulation in a randomly chosen position inside the grid of columns. 
In AW state the average firing rate is about 8.8Hz with about 1125 synapses per neuron. Not surprisingly, NEST started to converge to the production of stable Power Spectral Densities only when the integration time step was reduced down to 0.1ms. We used this time step for all simulations used for comparison with DPSNN.

\subsection{Execution Platforms}
\label{sec:platform}
The benchmark measures assessing the DPSNN scalability in terms of run time, initialization time, and memory usage herein described, were performed on the Galileo server platform provided by the CINECA consortium. Galileo is a cluster of 516 IBM nodes, each of which includes 16-cores, distributed on two Intel Xeon Haswell E5-2630 v3 octa-core processors clocked at 2.40GHz. The nodes are interconnected through an InfiniBand network. Due to the specific configuration of the server platform, the maximum-allowed partition of the server includes 64 nodes. Hyperthreading is disabled on all cores, therefore the number of MPI processes launched during each run exactly matches the hardware cores, with a maximum of 1024 hardware cores (or equivalently MPI processes) available for any single run. In the present version of the DPSNN simulator, parameters cannot be changed at runtime, and dynamic equations are explicitly coded (i.e., no meta-description is available as an interface to the user).

Simulations used for validation and comparison between DPSNN and NEST were run on a smaller cluster of eight nodes interconnected through a Mellanox InfiniBand network; each node is a dual-socket server with a six-core Intel(R) Xeon(R) E5-2630 v2 CPU (clocked at 2.60GHz) per socket with HyperThreading enabled, so that each core is able to run two processes.

\section{Results}
\label{sec:results}
\subsection{Initialization Times}
\begin{figure}[!hbt]
\centering
\includegraphics[width=1.0\textwidth]{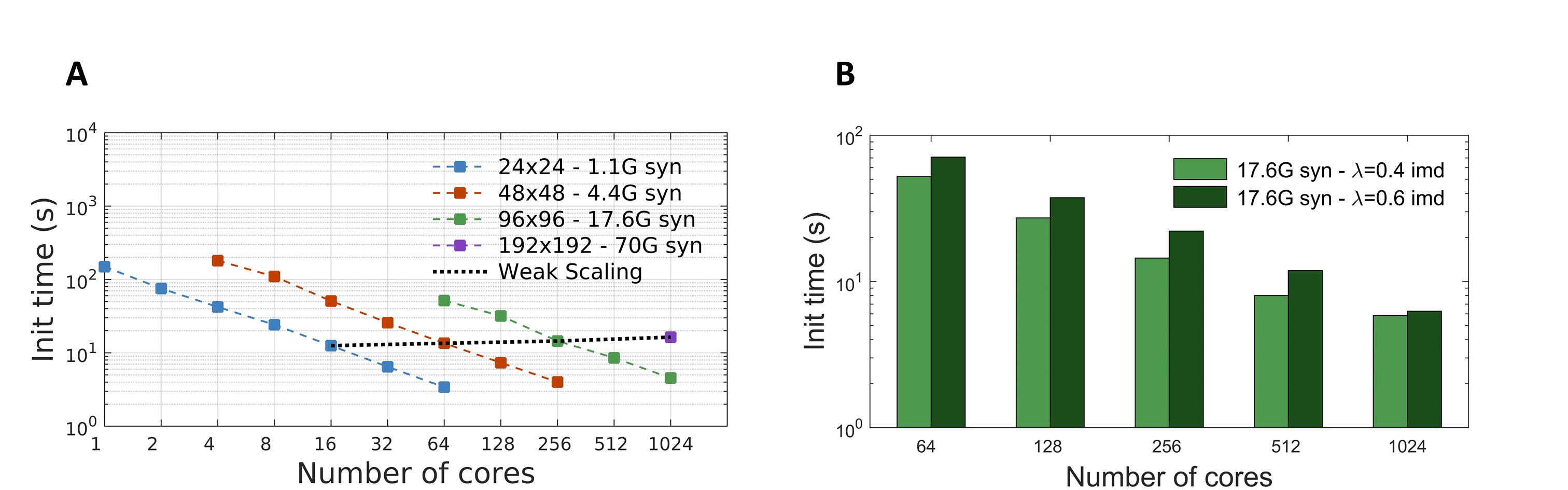}
\caption{Simulation initialization time. \textbf{(A)} Strong and weak scaling of the number of cores, for the four different network sizes described in Table \ref{tab:configs}. The value of $\lambda$ is kept constant, equal to 0.4 imd for the four networks. The nearly linear decrease in initialization time for fixed problem sizes indicates an efficient strong scaling, while the quality of weak scaling can be evaluated by observing the moderate increase of the initialization time for a four-fold increase in network size and hardware resources (black dotted line). \textbf{(B)} Initialization time scaling for a network of 17.6G synapses (96$\times$96 grid of columns) using two different $\lambda$ values: 0.4 imd(light green) and 0.6 imd (dark green). For higher connectivity ranges, an increase in the initialization time is registered.}
\label{fig:init}
\end{figure}

The initialization time, in seconds, is the time required to complete the setup phase of the simulator, which is necessary to build the whole neural network. Measures reported in Figure~\ref{fig:init}A show the scaling of the DPSNN initialization time for four network grid sizes, distributed over a growing number of hardware cores, which also correspond to MPI processes. Dashed colored lines represent strong scaling. Weak scaling is represented by the four points connected by the dotted black line (each point refers to a four-fold increase in the network size and number of used cores with respect to the previous one). For the explored network sizes and hardware resources, the initialization time speedup is almost ideal for fewer cores, while for the highest numbers of cores it is sub-ideal by a factor between 10\% and 20\%.

Figure~\ref{fig:init}B reports the scaling of the initialization time for two different values of the characteristic spatial scale of the decay of the connection probability ($\lambda$), for a single grid size (96$\times$96 columns). As already explained, for all simulations we kept the total number of synapses per neuron constant. 
Going from $\lambda=0.4$ to $\lambda=0.6$ imd, each excitatory neuron in a column has synaptic connections with neurons in a number of other columns growing from 44 to 78 (neglecting variations to proximity to columns' boundaries). Almost the same scaling is observed for both values of $\lambda$, with expected higher values for the initialization of a network with longer-range connectivity. Indeed, a dependence of the initialization time on synaptic connectivity range is expected, because it affects the proportion of target synapses residing in different columns, for which MPI messaging is needed for the connectivity setup, as explained in Materials and Methods.

\subsection{Memory Occupation}
In DPSNN, we expect the memory usage to be dominated by the representation of synapses, which are thousands per neuron, with only a minor impact of the representation of neurons and external synapses. Therefore, we defined a memory consumption metric as the total required memory divided by the number of recurrent synapses (see Section \ref{sec:perfMeasures}).

Each static synapse needs 12 bytes (see Table~\ref{tab:synrepresent}). As described in Section \ref{sec:simDesc}, during network initialization synapses are generated by the process storing the source neuron. Subsequently, their values are communicated to the process containing the target neuron. The result is that, only during the initialization phase, each synapse is represented at both source and target processes. Therefore, we expect a minimum of 24 bytes/(recurrent synapse) to be allocated during initialization, which is the moment of peak memory usage.

\begin{figure}[!hbt]
\centering
\includegraphics[width=1.0\textwidth]{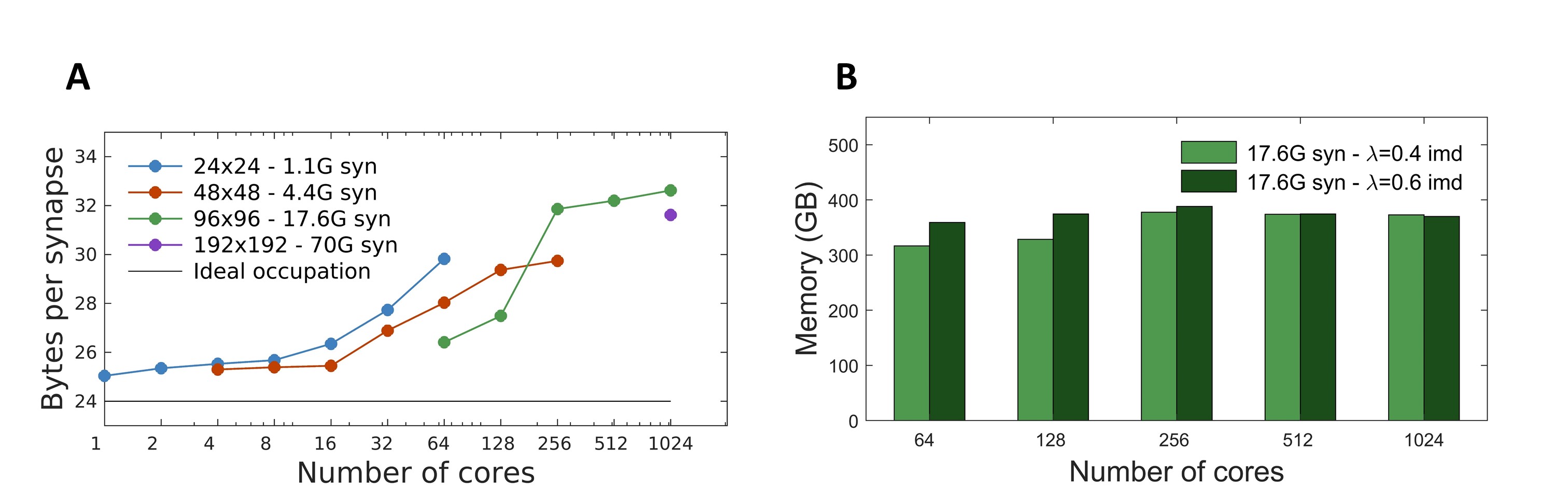}
\caption{Memory usage. \textbf{(A)} Memory footprint per synapse for four different networks sizes, distributed over different numbers of cores. Peak memory usage is observed at the end of network initialization, when each synapse is represented at both the source and target process, with a minimum expected cost of 24 byte/synapse (afterwards, memory is released on the source process). A different MPI overhead is observable below and over the threshold of single-node platforms (16 cores). \textbf{(B)} Memory footprint scaling with the number of cores for a network of 17.6G synapses (96 x 96 grid of columns) using two different $\lambda$ values: 0.4 imd (light green) and 0.6 imd(dark green). The total number of generated synapses is nearly the same and the memory footprint remains substantially constant.}
\label{fig:memUsage}
\end{figure}

The measured total memory consumption is between 25 and 32 bytes per recurrent synapse, for all simulations performed, from 1 to 1024 MPI processes (Figure~\ref{fig:memUsage}). From Figure~\ref{fig:memUsage}A we observe a different trend of the memory cost {\it vs} number of cores, below and above the threshold of single-node platforms (16 cores, see Section \ref{sec:platform}). Beyond a single node, additional memory is mainly required by the buffers allocated by MPI interconnect libraries that adopt different strategies for communications over shared memory (inside a node) or among multiple nodes (Figure~\ref{fig:memUsage}A). The memory occupation per synapse has been measured for different network sizes. Figure~\ref{fig:memUsage}A shows that, for a given number of cores, the memory overhead typically decreases for increasing size of the network, as expected.

Figure~\ref{fig:memUsage}B shows the impact of the spatial scale of connectivity decay on the memory footprint. Here the network has size 96$\times$96 columns (for a total of 17.6G synapses), and $\lambda=0.4, 0.6$ imd. As expected the memory footprint, mainly dependent on the total number of synapses,  essentially remains constant. Note that the relative differences for $\lambda=0.4, 0.6$ imd decrease as the number of cores increases; this is consistent with the fact that, given the network size, for larger core numbers more columns get distributed on processes residing in different cores. This dilutes the effect of $\lambda$, which goes in the same direction.

\subsection{Simulation Speed and its Scaling}
\subsubsection{Impact of Simulated State (SW or AW)}
\begin{figure}[!hbt]
\centering
\includegraphics[width=0.9\textwidth]{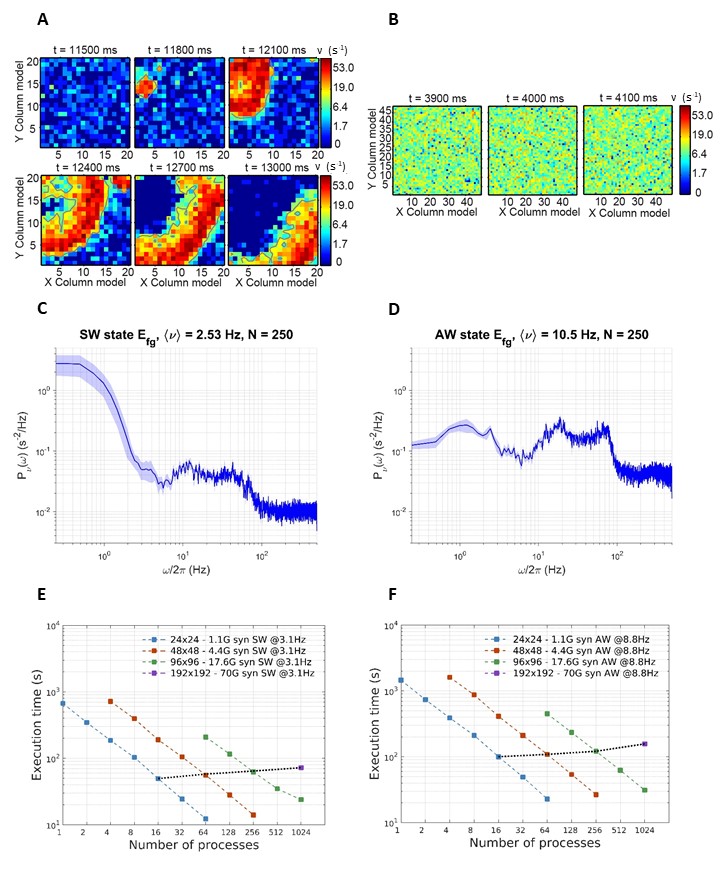}
\caption{Simulation time scaling and phenomenological behavior. \textbf{(A)} Time consecutive snapshots of the activity distribution in space, showing the wavefront propagation during a simulation expressing SW states. \textbf{(B)} Consecutive snapshots of the whole network activity in an asynchronous state. \textbf{(C-D)} Power spectra of network activity respectively in SW \textbf{(C)} and AW \textbf{(D)} states. \textbf{(E-F)} Scaling of wall-clock execution time for 1~s of SW \textbf{(E)} and AW \textbf{(F)} simulated activity. In both SW and AW states, the scaling has been measured on four different network sizes, as in Table \ref{tab:configs}.}
\label{fig:scaling}
\end{figure}

DPSNN execution times for SW and AW state simulations are reported in Figure~\ref{fig:scaling}. Figures~\ref{fig:scaling}A and ~\ref{fig:scaling}B show example snapshots of the simulated network activity in the SW and AW states, respectively; Figures~\ref{fig:scaling}C and~\ref{fig:scaling}D show corresponding power spectra of the network activity, confirming the main features predicted by theory for such states (such as the low-frequency power increase for SW, the high-frequency asymptote proportional to the average firing rate, the spectral resonances related to SFA, and delays of the recurrent synaptic interaction). Figures~\ref{fig:scaling}E and \ref{fig:scaling}F represent strong scaling for SW and AW: the wall-clock time required to simulate 1s of activity, {\it vs} number of processes, for four network sizes (see Table~\ref{tab:configs}). In the same plots, weak scaling behavior is measured by joining points referring to a four-fold increase in both the network size and the number of processes.

\begin{figure}[!hbt]
\centering
\includegraphics[width=1.0\textwidth]{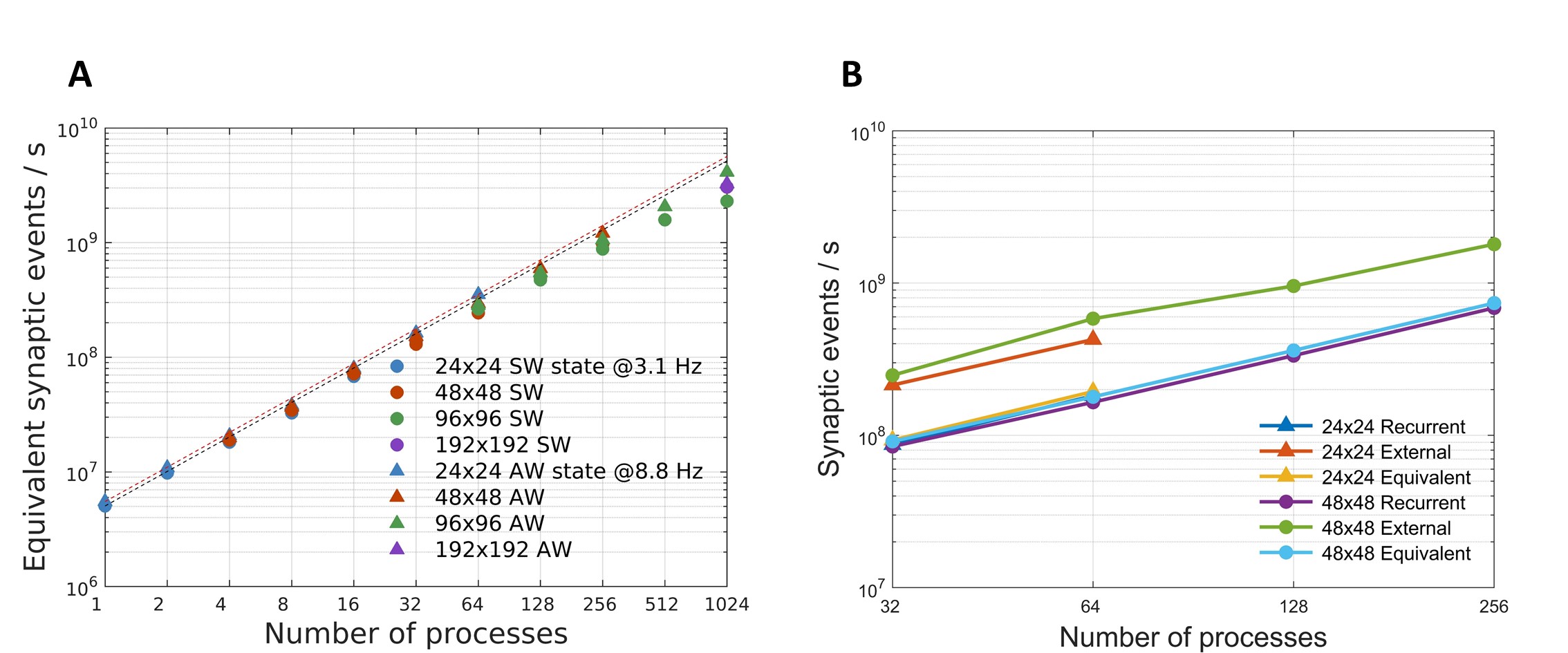}
\caption{Simulation speed-up. \textbf{(A)} Speed-up of the total number of equivalent simulated synaptic events per wall-clock second evaluated for SW (circles) and AW simulations (triangles) on the four used network configurations. Dashed lines stand for ideal speed-up of the simulated synaptic events for SW (black) and AW (red). \textbf{(B)} Scaling of recurrent, external and equivalent (recurrent + external) synaptic events per second for two configurations (24$\times$24 and 48$\times$48) in the AW state.}
\label{fig:synevents}
\end{figure}

Simulation speeds are measured using the equivalent synaptic-events-per-second metric, defined in Section \ref{sec:perfMeasures}. Departures from the ideal scaling behavior (linearly decreasing strong scaling plots, horizontal weak scaling plots) are globally captured in Figure~\ref{fig:synevents}A, for both SW and AW states and for all the problem sizes. 
For the simulations reported in this study, the simplified synaptic-events-per-second metric is a good approximation of a more complex metric separating recurrent and external events (Poisson noise), as demonstrated by Figure~\ref{fig:synevents}B: looking in more detail, the simulation of recurrent synaptic events is slower than that of external events that are locally generated by the routine that computes the dynamics of individual neurons. In our configurations there is about one order magnitude more recurrent synaptic event than external.

As numbers representative of the good scalability of the simulator, we remark that in the worst cases (1024 processes; 96$\times$96  for SW, 192$\times$192 for AW), the speedup with respect to the reference case (24$\times$24 on a single process) is about 590 for AW and 460 for SW, compared with the ideal speedup of 1024.

We also notice that performance scaling is slightly better for AW than SW, which is understood in terms of workload balance (due to the more homogeneous use of resources in AW); it is in any case remarkable that, although SW and AW dynamic states imply very different distributions of activity in space and time, the simulator provides comparable performance scaling figures.

Starting from the simulation speed expressed in terms of equivalent synaptic events per wall-clock second, DPSNN simulations presents a slow-down factor with respect to the real time, variable with the simulated network state and the available hw resources used for simulation. For example, considering a commodity cluster as the one used in this paper, able to allocate up to 64 processes, a network of size 24$\times$24 is about 12 times slower than real-time when simulating a SW state and about 23 times slower for the AW state. For larger simulations the slow-down factor increases, due to the not perfect scaling of the simulator performances. In this case, a 96$\times$96 network distributed on 1024 processes presents a slow-down factor of about 23 for a SW simulation and of about 31 for the AW state.

\subsubsection{Impact of Communication}

\begin{figure}[!hbt]
\centering
\includegraphics[width=1.0\textwidth]{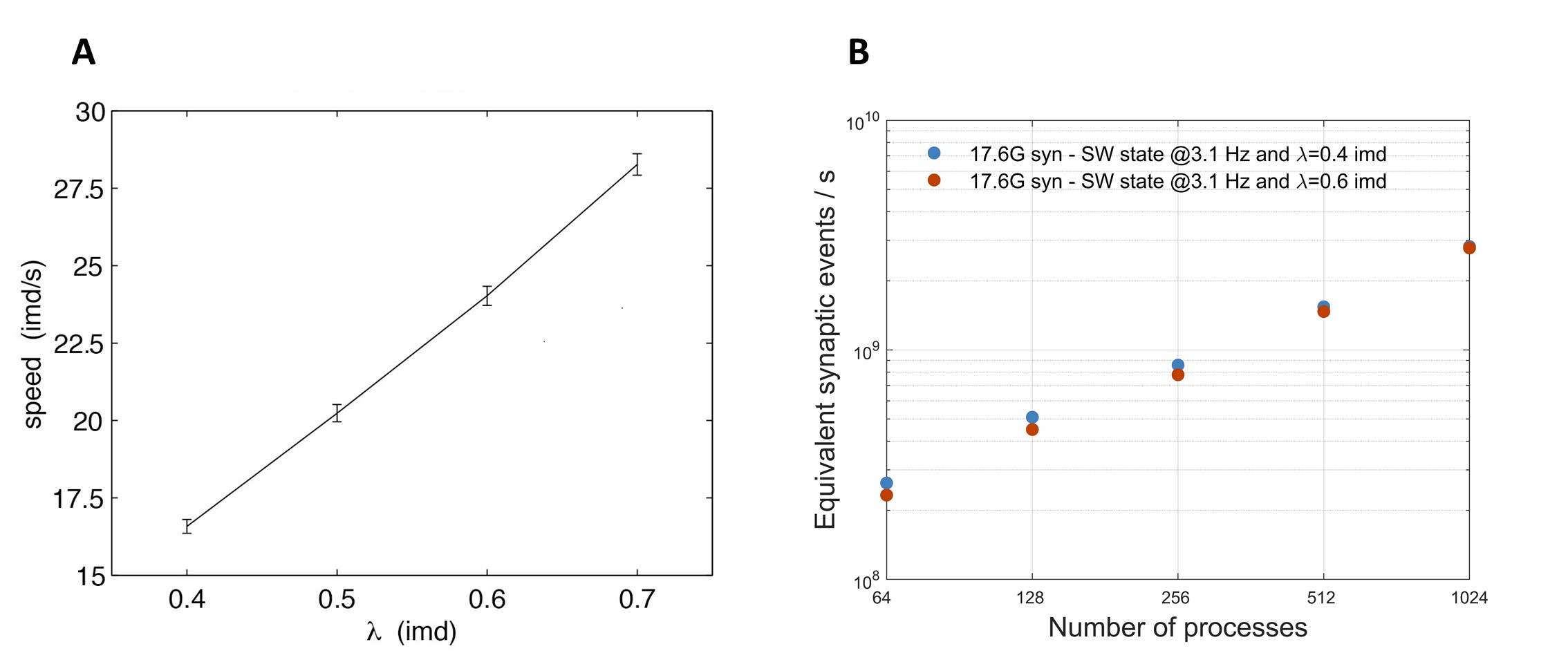}
\caption{Impact of the exponential connectivity parameter $\lambda$ on SW simulations. \textbf{(A)} Speed of waves generated by the SW model (3.1Hz) as a function of connectivity spatial scale $\lambda$. Average propagation speed of wavefronts relates to “wave activity” as a function of the parameter $\lambda$. For each point, the average is evaluated over 10 simulations, each time varying the connectivity matrix. In each simulation the average is computed over all observed waves. The vertical bars represent the standard deviation over different realizations of the connectivity. \textbf{(B)} scaling of SW simulations for two values of the exponential connectivity parameter $\lambda$: 0.4 imd (blue) and 0.6 imd (orange). The total number of generated synapses is nearly the same for both $\lambda$ values, with a different distribution of local and remote synapses, resulting in an increment of lateral connectivity for larger values, with almost no impact on performance. Scaling has been measured on a grid of 96$\times$96 columns, with 12M neurons and 17.6G synapses. Performance is expressed in terms of equivalent simulated synaptic events per wall clock second.}
\label{fig:impactSW}
\end{figure}

\cite{2019-PDP-RealTime} studied the relative impact of computation and communication on the performance of DPSNN applied to simulations of AW states. For neural network sizes and number of processes in the range of those reported in this paper, the time spent in computation is still dominant, while communication grows to be the dominant factor when the  number of neurons and synapses per process is reduced. In this paper, we evaluated the impact of communications on DPSNN performances both on SW and AW state simulations, using two different approaches. In the case of SW simulations, the analysis has been carried out with varying $\lambda$ (therefore varying the ratio of local versus remote excitatory synaptic connections, at a fixed total number of synapses per neuron):  $\lambda = 0.4$ imd (60 \% local connectivity), $\lambda = 0.6$ imd (35 \% local connectivity); clearly, higher $\lambda$ results in  higher payload in communication between processes.  Also, Figure~\ref{fig:impactSW}A shows the known linear dependence of the slow-wave speed on $\lambda$ \cite{coombes2005waves,capone:2017}. As the wave speed increases, the duration of the Up states stays approximately constant (not shown), so that an increasing portion of the network is simultaneously activated, which in turn may impact the simulation performance. In physical units, for an inter-modular distance (imd) of 0.4~mm, $\lambda = 0.6$ imd implies a spatial decay scale of 0.24 mm, and the corresponding speed is approximately 15~mm/s, which is in the range of biologically plausible values \cite{sanchez2000cellular,ruiz-mejias:2011,wester:2012,stroh2013making,capone:2019}.

\begin{figure}[!hbt]
\centering
\includegraphics[width=0.50\textwidth]{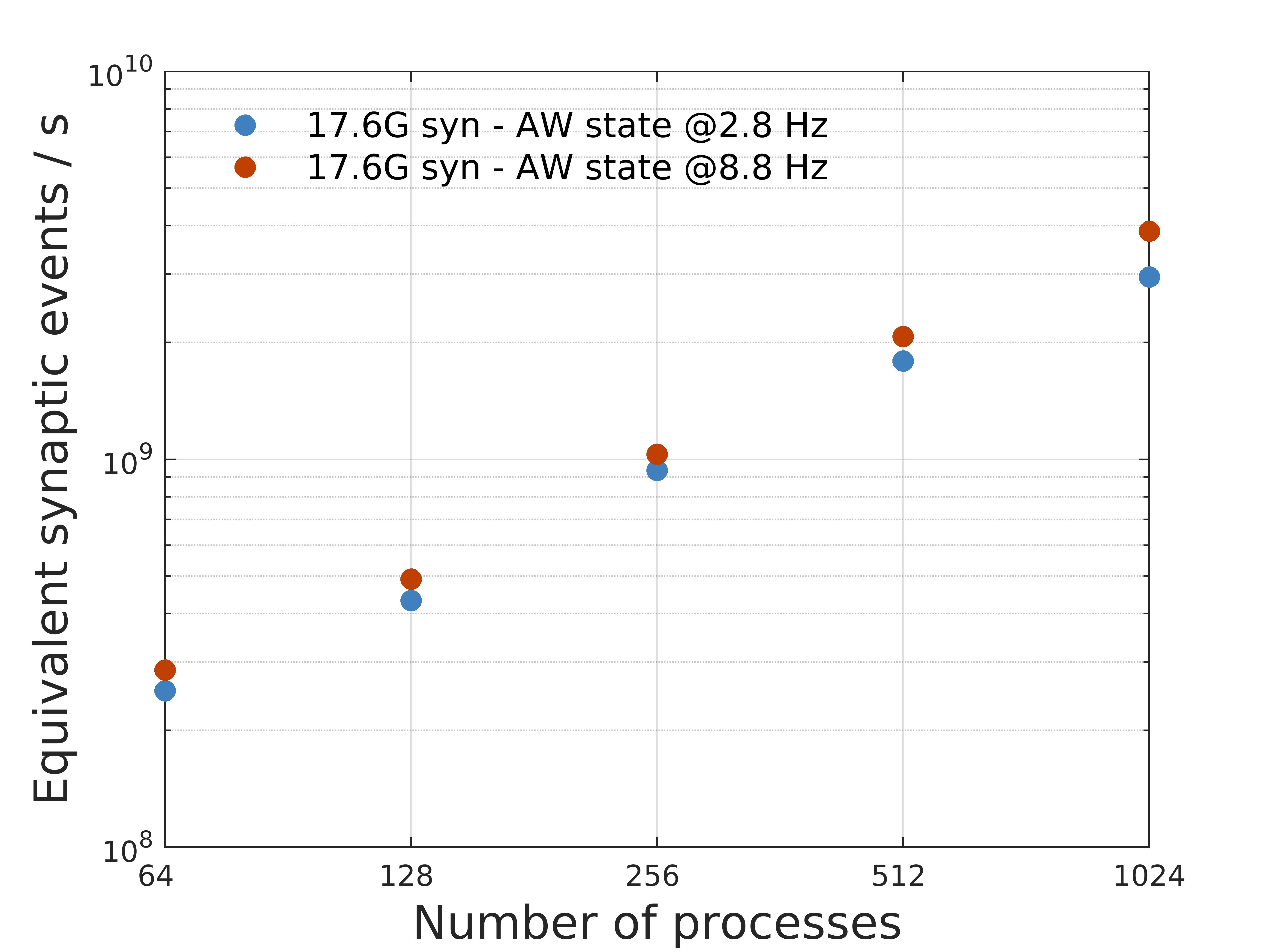}
\caption{Impact of different mean firing rates on AW state simulations. A network with 17.6G synapses (96$\times$96 grid of columns) has been simulated with two different values of mean firing rates, 2.8Hz (blue) and 8.8Hz (orange), corresponding to a different payload in communications. The plot shows a slight performance increase, in terms of equivalent simulated synaptic events per second, for systems with higher firing rate (and higher communication payload).}
\label{fig:impactAW}
\end{figure}

Figure~\ref{fig:impactSW}B shows that the impact of $\lambda$ on SW simulations is almost negligible. Figure~\ref{fig:impactAW} shows, instead, the impact of different mean firing rates on AW simulations. Higher firing rates imply higher payloads. DPSNN also demonstrates good scaling behavior in this case, with a slight performance increase for systems with a higher communication payload; this latter feature is due to communication costs being typically dominated by latencies and not bandwidth in spiking network simulations. 

\subsection{DPSNN and NEST: Comparison of Performances}
\label{sec:nest}
Table \ref{tab:NEST_DPSNN_comp} reports a performance comparison between DPSNN and NEST using the configuration, described in Section~\ref{sec:DPSNNvsNESTmaterial}. DPSNN is about 130 times faster than NEST for SW simulations and about 80 times faster for AW cases, and its initialization is about 19 times faster. The memory footprint of DPSNN is about 2.5 times lower due to the decisions of representing the identities of presynaptic and postsynaptic neurons with only 4 bytes per neuron and the storage of weights using only 2 bytes per synapse (see Table \ref{tab:synrepresent}). For the comparison of execution speed, we selected the best execution time for each simulator, for a fixed amount of used hardware resources; that is, the number of nodes of the cluster server. On the NEST simulator, we explored the space of all possible combination of MPI processes and number of threads during the AW simulations, in order to find the configuration performing better on a fixed number of hardware resources. In the same configuration, we also compared the initialization phase, the memory usage, and the SW execution times.

\begin{table}[htb]
\small
\caption{Memory footprint, initialization and execution times for 10~s of activity in SW and AW states required by DPSNN and NEST to simulate a neural network made of 2.8M neurons and 4.4G synapses (48$\times$48 grid of cortical columns). The second column reports the number of processes over which simulations are distributed. In case of DPSNN it corresponds to the number of pure MPI processes, while for NEST it is the number of Virtual Processes (VP). Each VP is calculated as MPI processes $\times$ number of threads, where six threads are used for all NEST simulations.}
\label{tab:NEST_DPSNN_comp}
\begin{center}
\begin{tabular}{|c|c|c|c|c|c|c|c|c|c|}
\hline
\multicolumn{2}{|c|}{\shortstack{Execution\\platform}} & \multicolumn{2}{|c|}{Init time (s)} & \multicolumn{2}{|c|}{SW Exec time (s)} & \multicolumn{2}{|c|}{AW Exec time (s)} & \multicolumn{2}{|c|}{Memory (GB)}\\
\hline
Nodes & \shortstack{Cores/\\Procs.} & DPSNN  & NEST &  DPSNN  & NEST & DPSNN  & NEST &  DPSNN  & NEST\\
\hline
2 & 48 & 29.2 & 603.4 & 1143.3 & 164555.3 & 2371.3 & 164008.8 & 78.8 & 208.0 \\
\hline
4 & 96 & 16.3 & 299.3 & 620.0 & 83305.0 & 1257.7 & 85025.2 & 78.4 & 200.8 \\
\hline
8 & 192 & 8.3 & 149.1 & 360.0 & 42601.7 & 684.9 & 42425.2 & 96.3 & 197.6 \\
\hline
\end{tabular}
\end{center}
\end{table}

\section{Discussion}
\label{sec:discussion}
We presented a parallel distributed neural simulator, with emphasis on the robustness of its performance and scaling with respect to quite different collective  dynamical regimes. This mixed time- and event-driven simulation engine (Figure~\ref{fig:flow}) has been used to simulate large-scale networks including up to 46 million point-like spiking neurons interconnected by 70 billion instantaneous current synapses.

The development of DPSNN originated from the need for a simple, yet representative benchmark (i.e. a mini-application) developed to support the hardware/software co-design of distributed and parallel neural simulators. Early versions of  DPSNN \cite{paolucci_dpsnn:2013} have been used to drive the development of the EURETILE system \cite{euretile_jsa:2016} in which a custom parallelization environment and the APEnet hardware interconnect were tested. DPSNN was then extended to incorporate the event-driven approach of \cite{mattia:2000}, implementing a mixed time-driven and event-driven strategy inspired by \cite{morrison:2005}. This simulator version is also currently included among the mini-applications driving the development of the interconnect system of the EXANEST ARM-based HPC architecture \cite{Katevenis:2016}.
In the framework of the Human Brain Project (https://www.humanbrainproject.eu), DPSNN is used to develop high-speed simulation of SW and AW states in multi-modular neural architectures. The modularity results from the organization of the network into densely connected modules mimicking the known modular structure of cortex.  In this modeling framework, the inter-modular synaptic connectivity decays exponentially with the distance.

In this section, we first discuss the main strengths of the simulation engine and then put our work into perspective by comparing the utility and performances of such a specialized engine with those of a widely used general-purpose neural simulator (NEST). 

\subsection{Speed, Scaling and Memory Footprint at Realistic Neural and Synaptic Densities}
DPSNN is a high-speed simulator. The speed ranged from $3\times10^{9}$ to $4.1\times10^{9}$ equivalent synaptic events per wall-clock second, depending on the network state and the connectivity range, on commodity clusters including up to 1024 hardware cores. As an order of magnitude, the simulation of a square cortical centimeter ($\sim 27\times10^{9}$ synapses) at realistic neural and synaptic densities is about 30 times slower than real time on 1024 cores (Figure~\ref{fig:synevents}).

DPSNN is memory parsimonious: the memory required for the above square cortical centimeter is 837~GB (31~byte/synapses, including all library overheads), which is in the range of commodity clusters with few nodes. The total memory consumption ranges between 25 and 32~byte per recurrent synapse for the whole set of simulated neural networks and all configurations of the execution platform (from 1 to 1024~MPI processes, Figure~\ref{fig:memUsage}). The choice of representing with 4 bytes the identities of presynaptic and postsynaptic neurons limits the total number of neurons in the network to 2 Billions. However, this is not yet a serious limitation for neural networks including thousands of synapses per neuron to be simulated on execution platforms including few hundreds of multi-core nodes. The size of neural ID representation will have to be enlarged for simulation of systems at the scale of human brains. Concerning synaptic weights, as already discussed, static synapses are stored with only two bytes of precision, but injection of current and neural dynamics is performed with double precision arithmetic. When plasticity is turned on, single precision floating-point storage of  LTP and LTD contributions is adopted.

The engine has very low initialization times. DPSNN takes about 4~s to set up a network with $\sim 17\times10^{9}$ synapses (Figure~\ref{fig:init}). We note that the initialization time is relevant, especially when many relatively short simulations are needed to explore a large parameter space. 

Its performance is robust: good weak and strong scaling have been measured in the observed range of hardware resources and for all sizes of simulated cortical grids. The simulation speed was nearly independent from the mean firing rate (Figure~\ref{fig:impactAW}), the range of connectivity (Figure~\ref{fig:impactSW}), and from the cortical dynamic state (AW/SW) (Figure~\ref{fig:synevents}). 

\subsection{Key Design Guidelines of the Simulation Engine}
A few design guidelines contribute to the speed and scaling of the simulation engine. We kept as driving criteria the goals of increasing the locality in memory accesses, the reduction of interprocess communication, an ordered traversal of lists and a reduction of backward searches and random accesses, the minimization of the number of layers in the calling stack, and a complete distribution of computation and storage among the cooperating processes with no need for centralized structures.  

We stored in the memory of a process a set of spatially neighbouring neurons, incoming synapses, and outgoing axons. For the kind of spatially organized neural networks described in this paper, this reduces the size of the payload and the required number of interprocess communications. Indeed, many spiking events will need to reach target neurons (and synapses)  stored in the same process of the spiking neuron itself. 

An explicit ordering in memory is adopted for outgoing and ingoing communication channels (representing the first branches in an axonal like arborizations). Explicit ordering in memory is also adopted for other data structures, like:  the lists of incoming spiking events and recurrent synapses, the set of incoming event queues associated to different synaptic delays. Lists are implemented as arrays, without internal pointers. Also explicit ordering is used for the set of target neurons. Neurons are progressively numbered with lower bits in their identifiers associated to their local identity and higher bits conveying the id of the hosting process. The explicit ordering of neurons and synapses reduces the time spent during the sequence of memory accesses that will be followed during the simulation steps. In particular, lists are travelled only once per communication step, e.g. first searching for target synapses that are targets of spiking events and then through an ordered list of target neurons. 

A third design criteria has been to keep the stack of nested function calls short, following the scheme described in Figure~\ref{fig:flow}, and attempting to group at each level multiple calls to computational methods accessing the same memory structure. 
As an example, both the generation of external synaptic events (e.g. poissonian stimulus) and the temporal reordering of recurrent and external synaptic events targeting a specific neuron, is deferred to the computation of the individual neural dynamics. The execution of this routine is supported by local queues storing all the synaptic currents targeting the individual neuron. This queue of events is accessed during a single call of the routine computing the dynamics of the neuron. In this case the data structure supports both locality in memory and in computation.

Similar design guidelines would be problematic to follow for general purpose simulation engines that are supposed to support maximum flexibility in the description and simulation of the data structures and of the dynamics of neurons, axonal arborizations and synapses. Moreover, higher abstraction requires separating the functionality of the simulation engine into independent modules. This would dictate a higher number of layers in the calling stack, more complex interfaces and data structures that hide details like their internal memory ordering.

\subsection{Comparison with a State-of-the-art User-friendly Simulator and Motivations for Specialized Engines}
There is a widely felt need for versatile, general-purpose neural simulators that offer a user-friendly interface for designing complex numerical experiments and provide the user with a wide set of models of proven scientific value. This boosted a number of initiatives (notably the NEST initiative, now central to the European Human Brain Project). However, such flexibility comes at a price. Performance-oriented engines, missing all the layers required for offering user generality and flexibility, contain the bare minimum code. In the case of DPSNN, this resulted in higher simulation speed, reduced memory footprint, and diminished initialization times (see Section~\ref{sec:nest} and Table~\ref{tab:NEST_DPSNN_comp}). In addition, optimization techniques developed for such engines on use-cases of proven scientific value can offer a template for future releases of general-purpose simulators. Indeed, this is what is happening in the current framework of cooperation with the NEST development team. Finally, engines stripped down to essential kernels constitute more easily manageable mini-application benchmarks for the hardware/software co-design of specialized simulation systems, because of easier profiling and simplified customizations on system software environments and hardware targets under development. 

\subsection{Future Works}
The present implementation of DPSNN demonstrated to be efficient for homogeneous bidimensional grids of neural columns and for their mapping of up to 1024 processes, and this facilitates a set of interesting scientific applications. However, further optimization could improve DPSNN performance, either in the perspective of moving simulations toward million-core exascale platforms or targeting real-time simulations at smaller scale (\cite{2019-PDP-RealTime}), in particular addressing sleep-induced optimizations in cognitive tasks like classification (\cite{2019-SleepAndMemory}). For instance, we expect that the delivery of spiking messages will be a key element to be further optimized (e.g., using a hierarchical communication strategy). This will also be beneficial for an efficient support of white-matter long-range connectivity (brain connectomes) between multiple cortical areas.

\section*{Acknowledgments}
This research has received funding from the European Union’s Horizon 2020 Framework Programme for Research and Innovation under the Specific Grant Agreements No. 785907 (Human Brain Project SGA2), No. 720270 (HBP SGA1) and No. 671553 (EXANEST).
Large-scale simulations have been performed on the Galileo platform, provided by the CINECA in the frameworks of HBP SGA and of the INFN-CINECA collaboration on the “Computational theoretical physics initiative”.
We acknowledge Hans Ekkehard Plesser and Dimitri Plotnikov for their support in setting up NEST simulations. 
We are grateful to the members of the INFN APE Parallel/Distributed Computing laboratory for their strenuous support. 

\bibliographystyle{ieeetr}

\section{Supplementary Material}
\subsection{Initial Construction of Connectivity Infrastructure}
\label{sec:supplInitConstruct}
During the initialization phase, each process contributes to an awareness about the subset of processes that should be listened to during subsequent simulation iterations. At the end of this construction phase, each “target” process should know about the subset of “source” processes that need to communicate with it, and should have created its database of locally incoming axons and synapses. A simple implementation of the construction phase can be realized using two steps.
During the first step, each source process informs other processes about the existence of incoming axons and about the number of incoming synapses to be established. A single word, the synapse counter, is communicated among pairs of processes. Under MPI, this can be achieved by an MPI\_Alltoall(). Performed once, and with a single-word payload, the cost of this first step creates a cumulative network load proportional to the square of the number of processes. The cost of this operation is negligible in the range of processes explored by this paper.
The second step transfers the identities of synapses to be created on each target process. Under MPI, the payload, a list of synapses specific for each pair in the subset of processes to be connected, can be transferred using a call to the MPI\_alltoallv() library function. The cumulative load created by this second step is proportional to the product of the total number of processes and the subset of target processes reached by each source process.
The first step produces two effects: (1) it reduces the cost of the initial construction of synapses, the second step of the construction phase; and (2) the knowledge about the non-existence of a connection between a pair of processes can be used to reduce the cost of spiking transmission during the simulation iterations.

\subsection{Delivery of Spiking Messages during the Simulation Phase}
\label{sec:supplSpikeDeli}
Here, we describe the present implementation of the delivery of spiking messages. In this first implementation, we did not take advantage of the possibility of delivering spikes to targets just before the deadline imposed by the synaptic-specific delay. Instead, we used a synchronous approach: all spikes are delivered to target processes before proceeding to the simulation of the next time iteration of the neural dynamic. The delivery of spiking messages can be split into two steps, with communications directed toward subsets of decreasing sizes. During the first step, single-word messages (spike counters) are sent to the subset of potentially connected target processes. On each pair of source-target process subsets, the individual spike counter gives information about the actual payload (i.e., axonal spikes) that will have to be delivered, or about the absence of spikes to be transmitted between the pair. The knowledge of the subset has been created during the first step of the initialization phase, described in a previous section. The second step uses the spiking counter information to establish a communication channel solely between pairs of processes that actually need to transfer an axonal spike payload during the current simulation time iteration. In MPI, both steps can be implemented using calls to the MPI\_Alltoallv() library function. However the two calls establish actual channels among sets of processes of decreasing size, as described previously.

\subsection{Simulation Parameters}
\label{sec:supplSimParams}
The connectivity of a single module can be fully described by setting the values of recurrent synaptic efficacies ($J_{ts}^0$)  and of external stimuli ($J_{t,ext}^0$ ). According to mean-field theory, specific values must be set depending on the network state that has to be simulated. Table \ref{tab:parameters} summarizes the values of synaptic efficacies, both for recurrent and external connectivity and for each simulated state.

The dynamic of the LIF neurons with SFA, used throughout all the simulations reported in this paper, is described by equation 2 in the main article. The values of all the parameters are summarized in Table \ref{tab:parameters}.

\begin{table*}[hbt]
\caption{Simulation parameters. A summary of recurrent synaptic efficacies, external stimulus, and neural dynamics parameters. Recurrent synapses: $J_{t,s}$ are the values of synapses connecting the target neuron \textit{t} with the source neuron \textit{s}, for each simulated state. External stimulus: $J_{t,ext}$ represents the mean value of the efficacy of external synapses afferent the target neuron \textit{t}; $\nu_{t,ext}$ represents its mean firing rate; $N_{t,ext}$ is the number of external Poissonian trains per target neuron\textit{ t} for each simulated state. Neural dynamics parameters: excitatory and inhibitory neurons are modeled according to equation 2 in the main article. Inhibitory neurons have no adaptation, therefore the second equation does not apply.}
\label{tab:parameters}
\small
\begin{center}
\begin{tabular}{|c|c|c|c|c|c|c|c|c|c|}
\hline
\multicolumn{10}{|c|}{Recurrent synapses} \rule{0pt}{10pt}\\
\hline
State & $J_{F,F}$ & $J_{B,F}$ & $J_{I,F}$ & $J_{F,B}$ & $J_{B,B}$ & $J_{I,B}$ & $J_{F,I}$ & $J_{B,I}$ & $J_{I,I}$\\
\hline
SW 3.1 Hz & 0.600 & 0.382 & 0.560 & 0.382 & 0.429 & 0.560 & 3.17 & 3.17 & 3.0\\
\hline
AW 2.8 Hz & 0.515 & 0.412 & 0.560 & 0.412 & 0.429 & 0.560 & -1.5 & -1.5 & -1.5\\
\hline
AW 8.8 Hz & 0.515 & 0.412 & 0.560 & 0.412 & 0.429 & 0.560 & -1.5 & -1.5 & -1.5\\
\hline
\hline
\multicolumn{10}{|c|}{External stimulus} \rule{0pt}{10pt}\\
\hline
State & $J_{F,ext}$ & $J_{B,ext}$ & $J_{I,ext}$ & $\nu_{F,ext}$ & $\nu_{B,ext}$ & $\nu_{I,ext}$ & $N_{F,ext}$ & $N_{B,ext}$ & $N_{I,ext}$ \\
\hline
SW 3.1 Hz & 0.832 & 0.858 & 1.120 & 3.17 & 3.17 & 3.0 & 400 & 400 & 400 \\
\hline
AW 2.8 Hz & 0.858 & 0.858 & 1.120 & 3.17 & 3.17 & 3.0 & 400 & 400 & 400 \\
\hline
AW 8.8 Hz & 1.416 & 1.416 & 1.120 & 3.17 & 3.17 & 3.0 & 400 & 400 & 400 \\
\hline
\hline
\multicolumn{10}{|c|}{Neural dynamics parameters} \rule{0pt}{10pt}\\
\hline
\shortstack{Neural \\ kind} & \shortstack{$\tau_{m}$ \\ (ms)} & \shortstack{$C_{m}$ \\ (pF)} & \shortstack{E \\ (mV)} & \shortstack{$V_{\theta}$ \\ (mV)} & \shortstack{$V_{r}$ \\ (mV)} & \shortstack{$\tau_{arp}$ \\ (ms)} & \shortstack{$\alpha_{w}$ \\ (mV)} & \shortstack{$\tau_{w}$ \\ (ms)} & \shortstack{$g_{w}$ \\ (nS)} \\
\hline
Exc & 20 & 1 & 0 & 20 & 15 & 2 & 1 & 1000 & 0.02\\
\hline
Inh & 10 & 1 & 0 & 20 & 15 & 1 & - & - & -\\
\hline
\end{tabular}
\end{center}
\end{table*}

\end{document}